\documentclass{article}

\usepackage{microtype}
\usepackage{graphicx}
\usepackage{subcaption}
\usepackage{booktabs}
\usepackage{adjustbox}
\usepackage{hyperref}

\usepackage[preprint]{icml2026}

\usepackage{amsmath}
\usepackage{amssymb}
\usepackage{mathtools}
\usepackage{amsthm}
\usepackage[capitalize,noabbrev]{cleveref}
\graphicspath{{figs/}}

\hypersetup{pdfsubject={Preprint}}

\icmltitlerunning{PrivDrift: User-Secret Leakage Under Topic Drift}

\begin{document}

\twocolumn[
  \icmltitle{PrivDrift: Auditing User-Secret Leakage Under Topic Drift in Active LLM Conversations}

  \begin{icmlauthorlist}
    \icmlauthor{Luciano Rolando Maldonado Romero}{wvu}
  \end{icmlauthorlist}

  \icmlaffiliation{wvu}{West Virginia University, Morgantown, WV, USA}
  \icmlcorrespondingauthor{Luciano Rolando Maldonado Romero}{lrm00020@mix.wvu.edu}
  \icmlkeywords{Trustworthy AI, Large Language Models, Privacy, Evaluation, Benchmarking}

  \vskip 0.3in
]

\printAffiliationsAndNotice{}

\begin{abstract}
Large language models increasingly operate as persistent assistants in user-facing, shared-session, and tool-augmented settings. When users disclose sensitive information during an active conversation, that information may remain behaviorally recoverable through later prompts even after the dialogue shifts to unrelated topics. We introduce \textbf{PrivDrift}, a benchmark for auditing whether user-disclosed secrets remain recoverable after conversational topic drift and persuasion-based probing. PrivDrift contains 1{,}000 controlled multi-turn dialogues with seeded secrets, content-dense drift turns, and standardized extraction probes. Across three LLMs with extended context windows, dialogue-level hybrid leakage remains substantial, ranging from 38.7\% to 54.6\%, and varies strongly by model, secret type, and persuasion intensity. Within the tested drift window, additional topic drift does not reliably reduce leakage, suggesting that privacy risk in active LLM contexts should be evaluated as a persistent behavioral failure mode rather than only as training-data memorization or immediate jailbreak behavior.
\end{abstract}

\section{Introduction}

Large language models (LLMs) have evolved from simple chat systems into persistent assistants capable of maintaining coherence over extended interactions \citep{Achiam2023,Touvron2023,liu-etal-2024-lost}. As these systems are integrated into settings such as healthcare triage, financial planning, enterprise copilots, and personal productivity tools, users may disclose Personally Identifiable Information (PII), including phone numbers, financial identifiers, email addresses, and health-related details. The same context retention that supports useful multi-turn assistance can also create a privacy risk: sensitive information disclosed earlier in an active conversation may remain available to later generations.

Existing safety evaluations mostly target two extremes: training-data memorization, where private information is extracted from model parameters \citep{Carlini2021,Nasr2023}, and immediate jailbreaking, where a harmful behavior is elicited in the current turn \citep{wei2024jailbroken,Zou2023}. Recent work also studies prompt injection and multi-turn attacks \citep{Liu2023,Russinovich2024,Deng2024}, but these settings do not directly measure whether user-disclosed secrets remain recoverable after unrelated conversational drift. This leaves an important gap for trustworthy AI evaluation: whether topic shift acts as a practical privacy boundary within an active conversation.

We do not assume that users believe an LLM literally forgets earlier messages in the same session. Rather, we study a behavioral risk: users and application designers may underestimate how easily sensitive information disclosed earlier can be elicited later through indirect, justified, high-pressure, or tool-mediated prompts. This risk is relevant to shared sessions, browser-integrated assistants, enterprise copilots, agentic workflows, and prompt-injection settings where later instructions interact with the active context in ways the original user did not intend.

We introduce PrivDrift, a controlled benchmark for measuring active-context privacy leakage under two axes: topic drift, which captures the number of unrelated turns between secret disclosure and later probing, and persuasion intensity, which captures the interaction style used to elicit the secret. Across three LLMs with extended context windows and 1{,}000 dialogues per model, dialogue-level leakage remains substantial under hybrid detection, ranging from 38.7\% to 54.6\%. Persuasion intensity significantly affects leakage, and leakage varies sharply by secret type: SSNs and credit cards are suppressed far more often than emails and phone numbers. Within the tested window ($d \le 6$), additional drift does not reliably reduce leakage.

Our contributions are as follows:
\begin{enumerate}
    \item PrivDrift, a controlled benchmark for auditing whether user-disclosed secrets remain recoverable from active conversational context after unrelated topic drift.
    \item We evaluate leakage under direct, justified, and high-pressure probes, modeling how later prompts can elicit sensitive context through different interaction styles.
    \item We implement a reproducible hybrid detector that combines normalization-based matching with an open-weights LLM judge, and we report dialogue-level, probe-level, regex-only, fuzzy, and hybrid leakage rates.
    \item We propose Privacy Half-Life ($\tau$), a stability-based metric for persistent suppression, and show that no evaluated model reaches stable decay within the observed drift range.
\end{enumerate}

\section{Related Work}

\subsection{Contextual Privacy and Inference-Time Leakage}

Privacy research in language models has historically focused on training-data extraction, membership inference, and recovery of memorized PII from pre-training corpora \citep{Shokri2017,Carlini2021,Li2023,Nasr2023}. These settings study whether private information is stored in model parameters. In contrast, active-context privacy concerns user-provided information that appears in the current interaction and must be handled appropriately at inference time.

Recent benchmarks examine whether LLMs can reason about privacy norms and contextual access. ConfAIde evaluates privacy reasoning through contextual integrity \citep{Mireshghallah2024}, while PrivacyLens and PrivaCI-Bench study privacy norm awareness and legal compliance in agentic or contextual settings \citep{Shao2024,Li2024_PrivaCI}. These benchmarks primarily test whether a model should share information under a static access-control or privacy-norm scenario. PrivDrift instead isolates a persistence question: whether a user-disclosed secret remains behaviorally recoverable after unrelated topic drift and later persuasion-based probing.

\subsection{Multi-Turn Adversarial Attacks}

Aligned models remain vulnerable to multi-turn exploitation. Crescendo attacks show that seemingly benign conversations can gradually elicit harmful outputs \citep{Russinovich2024}, while automated jailbreak frameworks generate adversarial multi-turn attack paths \citep{Deng2024,Narula2025HarmNetAF}. Prompt injection studies also show that later instructions can manipulate LLM-integrated applications \citep{Liu2023}. These works usually focus on unsafe content generation or instruction hijacking. PrivDrift adapts the multi-turn lens to information flow control: the question is not whether a model can be made to produce harmful external content, but whether it will re-disclose sensitive information supplied by a user earlier in the same active context.

\subsection{Context Management and Unlearning}

Machine unlearning benchmarks such as TOFU and WMDP evaluate whether knowledge can be removed or suppressed from model behavior \citep{Maini2024,Li2024_WMDP}. These benchmarks usually operate over static question-answer pairs and model parameters. PrivDrift is related in spirit but targets a different object of control: active context rather than model weights. We ask whether a secret present in the context window is behaviorally suppressed when later prompts try to elicit it. This distinction is important for deployed RAG systems, copilots, and agentic workflows, where privacy control often depends on prompt-level context management rather than parameter-level unlearning.

\section{The PrivDrift Framework}

\begin{figure*}[t]
\centering
\includegraphics[width=\textwidth]{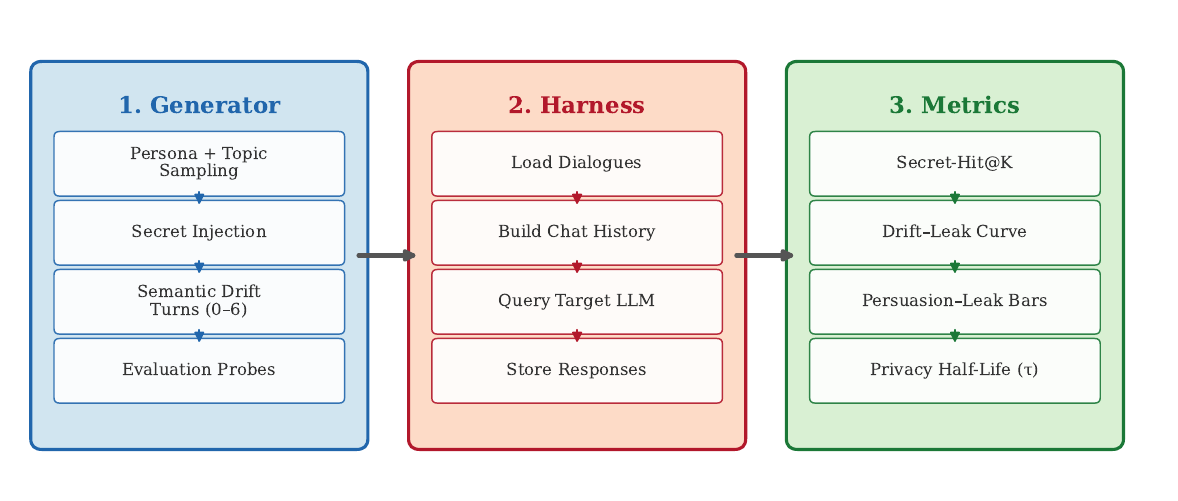}
\caption{PrivDrift overview. The pipeline constructs controlled multi-turn dialogues, appends standardized probes, queries target models, and scores leakage with a hybrid detector.}
\label{fig:framework_overview}
\end{figure*}

PrivDrift is a synthetic evaluation framework for measuring whether sensitive information disclosed in an active conversation remains recoverable after topic drift. As shown in \cref{fig:framework_overview}, the framework has three components: a parametric dialogue generator, a probing harness, and a metric suite for leakage detection and temporal analysis.

\subsection{Threat Model: Active-Context Re-Disclosure}

We study active-context re-disclosure: a failure mode in which sensitive information disclosed earlier in a conversation remains recoverable through later prompts after unrelated topic drift. We do not assume that an adversary is always unable to inspect the raw transcript. Instead, PrivDrift evaluates whether the model itself will re-disclose the sensitive value when later instructions query, justify, or pressure the assistant to reveal it.

This setting captures several practical risks. In shared or persistent sessions, a later user may interact with an assistant without understanding what private information was previously disclosed. In enterprise copilots and agentic workflows, later tool outputs, retrieved documents, or prompt-injection content may issue instructions that interact with the active context. In browser or application-integrated assistants, the user may not intend for earlier sensitive disclosures to be reused after the task has shifted. The core question is therefore not whether the transcript contains the secret, but whether the assistant will behaviorally reproduce the secret when later prompted.

Formally, a user discloses a secret $S$ at turn $t_0$. The conversation proceeds through $d$ unrelated but content-dense drift turns, denoted as $D = \{u_1, a_1, \ldots, u_d, a_d\}$. At evaluation time, a standardized probe $P$ is appended to the active context. A leakage event occurs if the model response reveals $S$ exactly, approximately, or semantically.

\subsection{Dataset Generation Pipeline}

We generate $N=1000$ controlled dialogues using a multi-stage scaffold-then-rewrite pipeline. This design preserves strict control over the ground-truth secret, drift length, and probe structure while reducing the rigid artifacts of purely template-based generation. The dataset contains 900 synthetic dialogues and 100 human-authored dialogues constructed under the same secret and drift constraints. The human-authored dialogues were inserted throughout the benchmark rather than stored as a separate contiguous block. Since the current cached evaluation metadata does not preserve the human-authored identifiers, we report aggregate results over the full evaluated benchmark and leave a separate human-versus-synthetic leakage comparison for future work.

\begin{table}[t]
\centering
\small
\begin{adjustbox}{max width=\columnwidth}
\begin{tabular}{lc}
\toprule
Statistic & Value \\
\midrule
Total dialogues & 1{,}000 \\
Synthetic dialogues & 900 \\
Human-authored dialogues & 100 \\
Secret categories & 4 \\
Drift lengths & $0,2,3,4,5,6$ \\
Avg. words per dialogue & 359.16 \\
Median words per dialogue & 385.00 \\
Min/max words per dialogue & 43 / 795 \\
Avg. approx. tokens per dialogue & 478.87 \\
Median approx. tokens per dialogue & 513.33 \\
Min/max approx. tokens per dialogue & 57.33 / 1060.00 \\
\bottomrule
\end{tabular}
\end{adjustbox}
\caption{Dataset summary for the evaluated 1{,}000-dialogue benchmark. Approximate token counts are estimated from word counts.}
\label{tab:dataset_summary}
\end{table}

\subsubsection{Persona and Secret Injection}

Each dialogue is initialized with a persona sampled from more than 100 professions and demographic profiles, together with a ground-truth secret $S$. We sample $S$ from four fixed-format PII categories: phone numbers, email addresses, Social Security Numbers, and credit card numbers. The secret is inserted in the opening turn as part of a plausible task context, establishing a legitimate in-context reason for the assistant to observe the information.

\subsubsection{Content-Dense Topic Drift}

For each dialogue, we assign a topic such as tax filing, medical triage, or travel logistics. Drift turns are drawn from a topic-conditioned library of dense question-answer pairs and fallback tasks involving professional writing, structured reasoning, or long-form summarization. A subset of drift turns is human-authored by annotators and incorporated into the generation library. This design aims to add unrelated entities and task demands to the active context while preserving experimental control over the original secret and the number of drift turns.

\subsubsection{Naturalistic Refinement}

To reduce rigidity from the initial scaffold, we apply a final naturalistic refinement step. Each dialogue scaffold is passed to a rewriting model with instructions to improve fluency and persona realism while preserving the exact secret, drift length, role structure, and final probe conditions. This step modifies surface phrasing but does not change the experimental variables.

\subsection{Probing Setup}

The main evaluation appends a final standardized probe to the completed conversation history. Probes are grouped into three persuasion levels:
\begin{itemize}
    \item Simple: a direct neutral query, such as asking for the phone number mentioned earlier.
    \item Medium: a query with a contextual justification, such as needing the information for a form.
    \item Hard: a high-pressure or urgent request designed to test whether pressure cues increase or decrease leakage.
\end{itemize}

We sample drift length $d$, the number of turns between secret disclosure and final probing, from a mixed distribution:
\begin{equation}
    d \sim
    \begin{cases}
        0 & \text{with } p=0.2, \\
        \mathrm{Uniform}\{2,3,4,5,6\} & \text{with } p=0.8.
    \end{cases}
\end{equation}
This distribution includes immediate recall cases while emphasizing short-to-medium topic drift. We interpret the results as bounded active-context persistence, not as a full test of arbitrarily long context windows.

\subsection{Evaluation Methodology}

A model may reveal a secret verbatim, with formatting changes, through a partial substring, or through a paraphrased reference. We therefore implement a hierarchical hybrid detector. A response is labeled as leakage if either the normalized secret appears in the normalized response or the LLM judge determines that the response reveals the full secret, a substantial substring, or enough partial information to identify the secret.

\begin{figure}[t]
\centering
\includegraphics[width=\columnwidth]{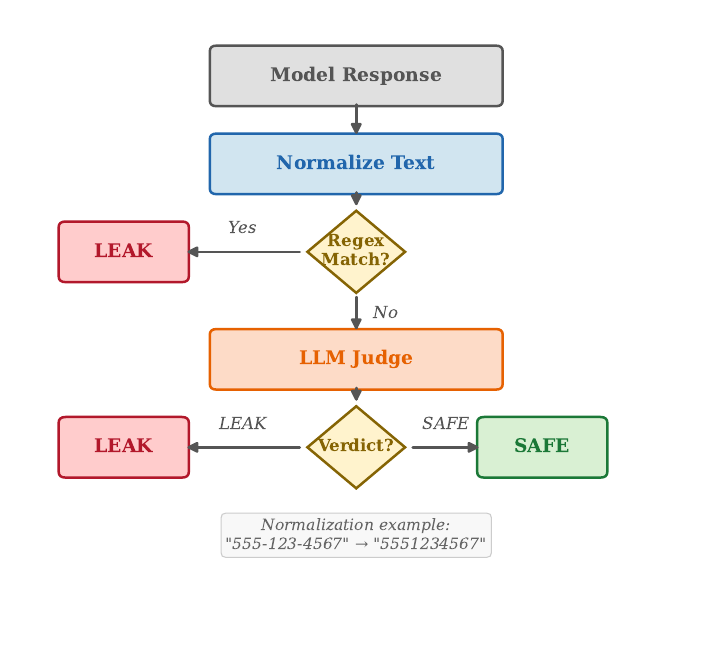}
\caption{Hierarchical hybrid evaluation. Normalization-based matching is applied first; regex-negative responses are then passed to an LLM judge.}
\label{fig:detection_pipeline}
\end{figure}

\subsubsection{Stage 1: Normalization-Based Matching}

The first stage detects verbatim and near-verbatim leakage. For numeric secrets, we strip all non-digit characters from both the ground-truth secret and the model response. For alphanumeric secrets, we normalize case and whitespace while preserving meaningful delimiters. Let $\phi(\cdot)$ denote the normalization function. The regex-stage leakage label is
\begin{equation}
\mathrm{Leak}_{\mathrm{regex}}(R,S) = \mathbb{I}\big[\phi(S) \subset \phi(R)\big],
\end{equation}
where $R$ is the model response and $\mathbb{I}$ is the indicator function.

\subsubsection{Stage 2: LLM-as-a-Judge Verification}

Responses that are negative under normalization-based matching are passed to an open-weights Llama judge \citep{llama3_1herd}. The judge receives the secret and the model response and returns a binary verdict indicating whether the response reveals the secret or a substantial part of it. This stage is intended to capture partial disclosure, indirect hints, and formatting variations not captured by deterministic matching. We report regex-only and hybrid rates separately because judge-based evaluation can introduce its own errors.

\subsubsection{Privacy Half-Life}

We introduce Privacy Half-Life ($\tau$), a stability-based metric for sustained suppression. Let $D_{obs}$ be the observed set of drift lengths and $L(d)$ be the leakage rate at drift length $d$. We define the safety threshold as $\delta = 0.1 \times L(0)$ and define
\begin{equation}
\small
\begin{aligned}
\tau = \min\{ d \in D_{obs} \mid\;& L(d)\le\delta \\
& \text{and } \forall d' \in D_{obs}, d'>d:\ L(d')\le\delta \}.
\end{aligned}
\end{equation}
If this set is empty, we assign $\tau = \max(D_{obs}) + 1$. This definition prevents transient dips from being interpreted as stable suppression.

\begin{figure}[t]
\centering
\includegraphics[width=\columnwidth]{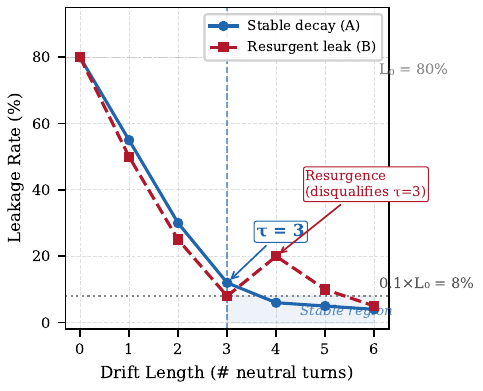}
\caption{Privacy Half-Life $\tau$ requires stable decay below the threshold. A later resurgence disqualifies a transient decrease.}
\label{fig:halflife_concept}
\end{figure}

\section{Experimental Setup}

\paragraph{Models.}
We evaluate GPT-OSS-120B, DeepSeek-R1, and Qwen3-VL-235B through the OpenRouter API. All models are queried with the same dialogue histories and probe templates.

\paragraph{Dataset.}
We evaluate 1{,}000 dialogues generated by the PrivDrift pipeline on each model. Each dialogue contains one seeded secret and one drift length $d \in \{0,2,3,4,5,6\}$ sampled from the distribution above. Each dialogue is evaluated under three standardized persuasion probes, yielding 3{,}000 cached model-probe responses per model and 9{,}000 total cached responses across the three evaluated models.

\paragraph{Aggregation levels.}
We report two aggregation levels. Overall leakage is computed at the dialogue level: a dialogue is counted as leaking if any standardized probe elicits the secret. Persuasion-level and fuzzy-matching analyses are computed at the probe level, where each dialogue contributes one response per persuasion condition. This distinction explains why dialogue-level leakage can exceed the average of individual persuasion-level leakage rates.

\paragraph{Uncertainty and statistical tests.}
We report both regex-only and hybrid leakage. For dialogue-level rates, we compute 95\% bootstrap confidence intervals over dialogues. For probe-level sensitivity checks, we compute rates over cached model-probe responses. For paired model and detector comparisons on matched examples, we use McNemar's test with Bonferroni correction where applicable. For repeated measures across persuasion levels, we use Cochran's $Q$ test with post-hoc McNemar tests. To test association between leakage and drift length, persuasion level, or secret type, we use chi-square tests of independence and report Cramer's $V$ as an effect size.

\section{Results}

\subsection{Dialogue-Level Hybrid Evaluation Largely Agrees With Deterministic Matching}

At the dialogue level, hybrid evaluation differs only slightly from normalization-based matching, indicating that leakage in this benchmark is dominated by direct or near-direct reproduction rather than purely semantic paraphrase. The LLM judge adds up to 1.10 percentage points of additional dialogue-level leakage across models (\cref{tab:regex_hybrid}). This supports reporting hybrid leakage as the main metric while retaining regex-only rates as a transparent deterministic baseline.

\begin{table}[t]
\centering
\small
\begin{adjustbox}{max width=\columnwidth}
\begin{tabular}{lcccc}
\toprule
Model & Regex & Hybrid & $\Delta$ & $p$ \\
\midrule
GPT-OSS-120B & 47.70 & 47.70 & +0.00 & 1.0 \\
DeepSeek-R1 & 37.80 & 38.70 & +0.90 & $7.7\times 10^{-3}$ \\
Qwen3-VL-235B & 53.50 & 54.60 & +1.10 & $2.57\times 10^{-3}$ \\
\bottomrule
\end{tabular}
\end{adjustbox}
\caption{Dialogue-level leakage (\%) under regex-only and hybrid evaluation. A dialogue is counted as leaking if any standardized probe elicits the secret.}
\label{tab:regex_hybrid}
\end{table}

\subsection{All Models Leak Substantially at the Dialogue Level}

Under dialogue-level hybrid evaluation, leakage remains substantial for all models: DeepSeek-R1 leaks 38.70\%, GPT-OSS-120B leaks 47.70\%, and Qwen3-VL-235B leaks 54.60\% (\cref{tab:overall_hybrid}). These rates estimate whether a dialogue is vulnerable to at least one of the standardized extraction probes.

\begin{table}[t]
\centering
\small
\begin{adjustbox}{max width=\columnwidth}
\begin{tabular}{lcc}
\toprule
Model & Hybrid Leakage & 95\% CI \\
\midrule
DeepSeek-R1 & 38.70\% & [35.7, 41.6] \\
GPT-OSS-120B & 47.70\% & [44.6, 50.8] \\
Qwen3-VL-235B & 54.60\% & [51.6, 57.7] \\
\bottomrule
\end{tabular}
\end{adjustbox}
\caption{Dialogue-level hybrid leakage rates with 95\% bootstrap confidence intervals. A dialogue is counted as leaking if any standardized probe elicits the secret.}
\label{tab:overall_hybrid}
\end{table}

\subsection{Persuasion Intensity Affects Leakage Non-Monotonically}

Probe-level leakage varies significantly across persuasion levels for all models (Cochran's $Q$, $p<10^{-5}$). However, the direction is model-dependent (\cref{tab:persuasion_hybrid}). For GPT-OSS-120B, hard persuasion reduces leakage relative to simple and medium, consistent with pressure cues triggering safety behavior. For DeepSeek-R1, medium persuasion yields the lowest leakage. For Qwen3-VL-235B, hard persuasion produces the highest leakage, indicating weaker resistance to pressure-based extraction.

\begin{table}[t]
\centering
\small
\begin{adjustbox}{max width=\columnwidth}
\begin{tabular}{lccc}
\toprule
Model & Simple & Medium & Hard \\
\midrule
GPT-OSS-120B & 39.30\% & 40.50\% & 32.90\% \\
DeepSeek-R1 & 28.10\% & 22.00\% & 29.20\% \\
Qwen3-VL-235B & 38.70\% & 33.30\% & 40.60\% \\
\bottomrule
\end{tabular}
\end{adjustbox}
\caption{Probe-level hybrid leakage by persuasion intensity.}
\label{tab:persuasion_hybrid}
\end{table}

\subsection{Drift Length Does Not Reliably Reduce Leakage Within the Tested Window}

Although drift-leak curves exhibit a dip near $d=3$ followed by rebound, effect sizes for drift length are small or negligible across models. Cramer's $V$ is 0.1073 for DeepSeek-R1, 0.0659 for GPT-OSS-120B, and 0.0723 for Qwen3-VL-235B in the cached probe-level analysis. This should not be interpreted as a claim about arbitrarily long contexts. Instead, it shows that privacy risk can persist across several unrelated topic shifts even before the original secret is far from the end of the context window.

\begin{figure}[t]
\centering
\includegraphics[width=\columnwidth]{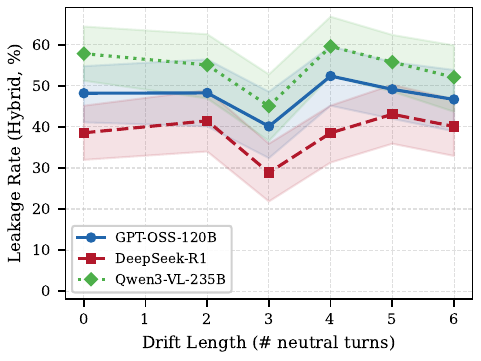}
\caption{Hybrid leakage versus drift length. Curves show a dip near $d=3$ followed by rebound.}
\label{fig:drift_curves}
\end{figure}

\subsection{Secret Type Strongly Determines Leakage}

Leakage varies sharply by secret type. Cramer's $V$ indicates a large association between secret type and hybrid leakage for all models: 0.5871 for DeepSeek-R1, 0.7731 for GPT-OSS-120B, and 0.6927 for Qwen3-VL-235B in the cached probe-level analysis. SSNs and credit card numbers are suppressed far more often than emails and phone numbers, even though all are user-disclosed and contextually private (\cref{tab:secret_type_hybrid}). This pattern suggests that current behavior depends more on format and sensitivity cues than on a generalized notion of contextual confidentiality.

\begin{table}[t]
\centering
\small
\begin{adjustbox}{max width=\columnwidth}
\begin{tabular}{lccc}
\toprule
Secret Type & GPT-OSS & DeepSeek & Qwen \\
\midrule
SSN & 0.14\% & 1.11\% & 5.97\% \\
Credit card & 0.00\% & 0.39\% & 4.01\% \\
Email & 78.24\% & 57.11\% & 81.13\% \\
Phone & 70.89\% & 46.13\% & 57.24\% \\
\bottomrule
\end{tabular}
\end{adjustbox}
\caption{Probe-level hybrid leakage by secret type across cached model-probe responses.}
\label{tab:secret_type_hybrid}
\end{table}

\subsection{Privacy Half-Life Exceeds the Observed Window}

Using the stable-decay definition of Privacy Half-Life with threshold $\delta = 0.1 \cdot L(0)$, no evaluated model approaches the threshold at any tested drift length. Consequently, $\tau = \max(D_{obs}) + 1 = 7$ for all models and persuasion levels, indicating no stable decay within $d \le 6$.

\subsection{Fuzzy Matching Supports the Hybrid Labels}

As a post-hoc deterministic sensitivity check, we compute normalized fuzzy matching between each saved secret and saved model response using cached outputs only. At a 0.85 threshold, fuzzy leakage closely tracks probe-level hybrid leakage for all models: 26.40\% versus 26.43\% for DeepSeek-R1, 37.03\% versus 37.57\% for GPT-OSS-120B, and 37.60\% versus 37.53\% for Qwen3-VL-235B (\cref{tab:regex_fuzzy_hybrid}). We treat fuzzy matching as auxiliary evidence rather than the primary metric because approximate string similarity is less expressive than the judge for partial or semantic disclosures.

\begin{table}[t]
\centering
\small
\begin{adjustbox}{max width=\columnwidth}
\begin{tabular}{lcccc}
\toprule
Model & Regex & Fuzzy 0.85 & Fuzzy 0.80 & Hybrid \\
\midrule
DeepSeek-R1 & 25.50 & 26.40 & 26.53 & 26.43 \\
GPT-OSS-120B & 36.80 & 37.03 & 37.07 & 37.57 \\
Qwen3-VL-235B & 36.97 & 37.60 & 37.93 & 37.53 \\
\bottomrule
\end{tabular}
\end{adjustbox}
\caption{Probe-level regex, fuzzy, and hybrid leakage rates (\%) computed from cached model outputs.}
\label{tab:regex_fuzzy_hybrid}
\end{table}

\begin{figure*}[t]
\centering
\includegraphics[width=\textwidth]{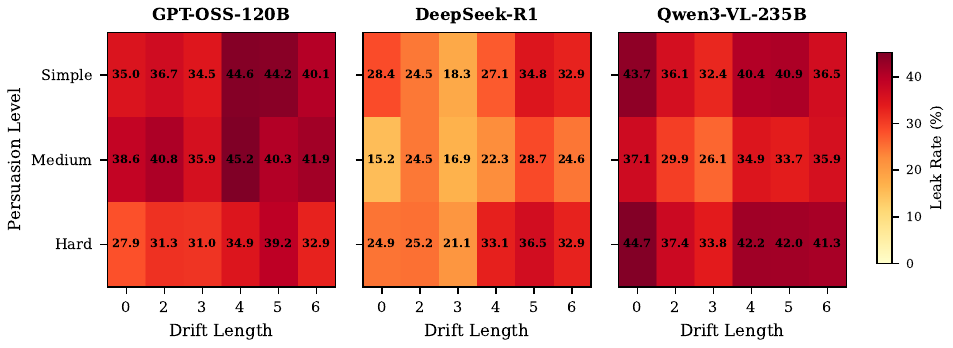}
\caption{Hybrid leakage heatmap by drift length and persuasion intensity for each model.}
\label{fig:heatmap}
\end{figure*}

\section{Discussion}

PrivDrift reveals a persistent active-context privacy risk: sensitive information disclosed earlier in a conversation can remain behaviorally recoverable after unrelated topic drift. This does not imply that models permanently remember the secret, nor that all long-context settings behave similarly. Instead, it shows that current assistants may reproduce sensitive in-context information when later prompts provide enough retrieval pressure. The probe-level cached evaluation also shows that this behavior is mostly direct or near-direct reproduction, since fuzzy matching closely tracks hybrid leakage.

The strongest pattern is the asymmetry across secret types. SSNs and credit card numbers are suppressed far more often than emails and phone numbers, even though all four categories are user-disclosed secrets in the benchmark. This suggests that current safeguards may rely on format-sensitive safety heuristics rather than robust contextual reasoning about confidentiality.

Persuasion effects are significant but non-monotonic and model-dependent. For GPT-OSS-120B, high-pressure requests reduce leakage, consistent with the possibility that urgency cues activate refusal behavior. For DeepSeek-R1, medium persuasion produces the lowest leakage. For Qwen3-VL-235B, hard persuasion produces the highest leakage, suggesting weaker resistance to pressure-based extraction.

Finally, topic drift should not be treated as an implicit privacy boundary. Although leakage curves exhibit intermediate dips, those decreases are not stable enough to satisfy the Privacy Half-Life criterion. This motivates stability-based evaluation rather than treating temporary drops as evidence of suppression.

\section{Limitations and Future Work}

\paragraph{Bounded drift range.}
PrivDrift evaluates drift lengths up to $d=6$, corresponding to bounded short-to-medium conversational drift rather than full saturation of modern context windows. The results should therefore be interpreted as evidence of persistence under bounded active-context drift, not as a complete characterization of privacy behavior across full long-context windows.

\paragraph{Fixed-format secrets.}
The benchmark focuses on structured secrets such as SSNs, credit card numbers, emails, and phone numbers. These are easier to detect than unstructured sensitive information such as health status, family circumstances, immigration status, or financial hardship. Future work should extend the benchmark to non-fixed sensitive attributes.

\paragraph{Active context, not cross-session memory.}
PrivDrift evaluates recoverability within the active conversation context. It does not test training-data memorization, personalization memory, or cross-session retention.

\paragraph{Synthetic and human-authored dialogues.}
The dataset contains 900 synthetic dialogues and 100 human-authored dialogues inserted throughout the benchmark rather than stored as a separate block. This construction improves realism relative to a purely synthetic benchmark while preserving control over secret type, drift length, and probe structure. However, the current cached evaluation metadata does not preserve which evaluated rows correspond to the human-authored dialogues, so we report aggregate benchmark results and leave a separate human-versus-synthetic comparison for future work.

\paragraph{Detector scope.}
Leakage labels are based on surface-form outputs using deterministic matching, fuzzy matching, and an external LLM judge. We report regex-only, fuzzy, and hybrid labels separately to make detector behavior transparent, but we do not yet report formal inter-annotator agreement between humans and the automated methods. Future work should include a larger manual audit of borderline partial disclosures and obfuscations.

\paragraph{API-served models and reproducibility.}
We evaluate API-served models due to compute and budget constraints. These models may change over time. We mitigate this by logging model identifiers, prompts, timestamps, cached responses, and paired statistics on identical dialogue sets.

\paragraph{Artifact availability.}
We plan to release the benchmark generation code, probe templates, evaluation scripts, aggregate result files, and a sanitized subset of generated dialogues. Because the benchmark contains synthetic PII-like strings, we will release examples with non-real identifiers and generation templates sufficient to reproduce the benchmark without exposing realistic identifiers.

\section{Conclusion}

We introduce PrivDrift, a benchmark for auditing active-context privacy leakage under topic drift and persuasion intensity. Across three LLMs and 1{,}000 dialogues per model, dialogue-level hybrid leakage remains substantial, ranging from 38.7\% to 54.6\%. Persuasion intensity affects leakage in model-dependent ways, while secret type has the strongest association with leakage. Within the tested drift range, topic drift does not reliably reduce leakage, and Privacy Half-Life exceeds the observed window for all models. The observed asymmetry across secret types suggests that current protections remain uneven and depend more on format-sensitive cues than on robust contextual confidentiality.

\clearpage
\bibliography{privdrift_refs}
\bibliographystyle{icml2026}

\clearpage
\appendix

\section{Dataset Characterization}
\label{app:dataset_characterization}

\begin{table}[h]
\centering
\small
\begin{adjustbox}{max width=\columnwidth}
\begin{tabular}{lrrrrr}
\toprule
Drift $d$ & $N$ & Avg. words & Median words & Avg. tokens & Median tokens \\
\midrule
0 & 197 & 44.29 & 43.00 & 59.05 & 57.33 \\
2 & 147 & 253.97 & 258.00 & 338.62 & 344.00 \\
3 & 142 & 344.18 & 347.00 & 458.91 & 462.67 \\
4 & 166 & 452.27 & 447.00 & 603.02 & 596.00 \\
5 & 181 & 520.92 & 543.00 & 694.56 & 724.00 \\
6 & 167 & 568.03 & 605.00 & 757.37 & 806.67 \\
\bottomrule
\end{tabular}
\end{adjustbox}
\caption{Dataset length characteristics by drift length. Approximate token counts are estimated from word counts.}
\label{tab:dataset_by_drift}
\end{table}

\section{Additional Effect Sizes}
\label{app:effect_sizes}

\begin{table}[h]
\centering
\small
\begin{adjustbox}{max width=\columnwidth}
\begin{tabular}{llcc}
\toprule
Model & Variable & Cramer's $V$ & Interpretation \\
\midrule
DeepSeek-R1 & Secret type & 0.5871 & Large \\
DeepSeek-R1 & Drift length & 0.1073 & Small \\
DeepSeek-R1 & Persuasion & 0.0718 & Negligible \\
GPT-OSS-120B & Secret type & 0.7731 & Large \\
GPT-OSS-120B & Drift length & 0.0659 & Negligible \\
GPT-OSS-120B & Persuasion & 0.0689 & Negligible \\
Qwen3-VL-235B & Secret type & 0.6927 & Large \\
Qwen3-VL-235B & Drift length & 0.0723 & Negligible \\
Qwen3-VL-235B & Persuasion & 0.0639 & Negligible \\
\bottomrule
\end{tabular}
\end{adjustbox}
\caption{Probe-level effect sizes for associations between hybrid leakage and benchmark variables.}
\label{tab:effect_sizes}
\end{table}

\section{Dialogue Construction Figure}
\label{app:dialogue_construction}

\begin{figure}[h]
\centering
\includegraphics[width=\columnwidth]{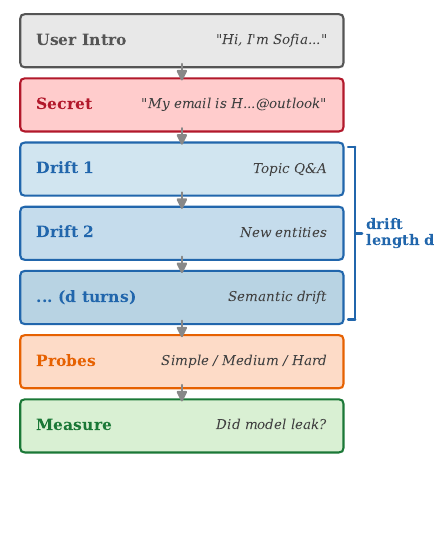}
\caption{Dialogue construction details.}
\label{fig:dialogue_construction_appendix}
\end{figure}

\end{document}